\documentclass[10pt]{article}
\usepackage[preprint]{tmlr}
\usepackage{amsmath,amssymb,amsthm,booktabs,graphicx,url}
\usepackage[hidelinks]{hyperref}
\newtheorem{proposition}{Proposition}
\newif\ifanon

\newcommand{\readerresult}{We then ran Qwen3.8-27B (pinned revision, think mode,
32,768-token cap, greedy) on 4 C2-changed inputs per network in arms C0, C2 and C4: 72
prompts. It passed the reader gate: C0 accuracy 0.875 against 0.344 with weights withheld
(difference 0.531, 95\% interval [0.354, 0.688]), with 5 of 72 answers unparseable (4 at
the token cap). On C2 it predicted the rewired class on 90.6\% of inputs and the original
class on 1.0\%: tracking minus stuck 0.896 [0.771, 1.000]. It tracked fully on 21
networks; both networks scoring 0 were parse failures. It spent about 25,000 tokens per
prompt on C0 and C2, against 1,400 with weights withheld. A C2-invariant rule that scores
0.948 on C0 was available to it, yet on C2 it behaved like X, not F. By the
pre-registered reading, on this population the reader's outputs track the rewired
function; this does not identify its internal strategy. This is one reader on
constructed 4-input networks; it does not say whether readers use fingerprints on trained
networks, where no eligible setting exists yet.}
\newcommand{\ladderslope}{$b=0.748$ (95\% CI $[0.620, 0.856]$; 10,000 bootstrap draws resampling networks)}

\anonfalse
\hypersetup{pdftitle={When Can a Rewiring Control Test LLM Weight Readers? C2 and an Eligibility Protocol},
  pdfauthor={Vaishak Girish Kumar and Sanika Vilas Najan}}

\title{When Can a Rewiring Control Test LLM Weight Readers?\\
C2 and an Eligibility Protocol}

\author{\name Vaishak Girish Kumar \email vaishakg@buffalo.edu \\
      \addr University at Buffalo
      \AND
      \name Sanika Vilas Najan \email snajan@buffalo.edu \\
      \addr University at Buffalo}

\begin{document}
\maketitle

\begin{abstract}
Large language models can predict the outputs of small neural networks from raw weights,
but success could reflect recognition of weight-space regularities (fingerprints) rather
than computation, and function-preserving controls cannot tell these apart. We propose
C2, a counterfactual that permutes the columns of a network's final linear readout. For
any network with a linear readout, C2 leaves every other parameter unchanged and
preserves the readout's singular values, norms and entry multiset, yet changes the
function in a way that can be recomputed exactly (Proposition~\ref{prop:c2}). We argue
that C2 separates fingerprinting from computation only under three conditions, which we
state as an eligibility protocol: a baseline restricted to C2-invariant summaries must
already predict the target on the original networks; C2 must change a recomputed
property that is specific to the rewiring; and the reader's identity must be attested.
No natural setting we tested is eligible. On 24 tiny MLPs a frontier reader route scored 1.000 on
original, permuted and rewired networks (0.292 with weights withheld) and followed the
rewired function on all 24 networks, but invariant-summary baselines were near chance
and the route's backend could not be attested. In two public zoos of 60 CNNs each, weight
statistics did not predict accuracy for held-out nonlinearity families. In-distribution,
they did on CIFAR-10 (mean absolute error 39\% below a constant and 13\% below a
family-mean baseline) but not on MNIST. On CIFAR-10, however, a random readout swap
predicted the C2 effect on accuracy (correlation 0.975); per-image targets made the effect
specific but left no working invariant strategy, so the conditions traded off. On a
constructed population where a planted fingerprint rule works per input, all conditions
hold and C2 separates a rule-following reference reader from an executor; an attested
open-weight reader (Qwen3.8-27B) followed the rewired function on 90.6\% of changed inputs
and the original on 1.0\%; with thinking disabled we found no evidence that it used the
weights. A registered replication on 24 fresh networks confirmed both
findings (tracking minus stuck 0.875, 95\% interval [0.750, 1.000]; no evidence of weight use
with thinking disabled), and a second reader family, DeepSeek-R1-Distill-Qwen-14B, also followed the
rewired function (tracking minus stuck 0.729 [0.552, 0.885]). The transforms, audits and raw
records are available from the authors.
\end{abstract}

\section{Introduction}
Language models can identify, describe, and sometimes edit neural networks given only
their weights~\citep{wsp,oracles}. Two explanations fit such results. A reader may
compute: it executes, approximately or exactly, the forward pass the weights define. Or it
may recognise fingerprints: regularities of trained weights, such as norms, spectra, or
value distributions, that correlate with what a network does. Several recent evaluations
probe readers with function-preserving transformations, such as hidden-unit permutations,
sign flips, and positive rescalings~\citep{huang,orbit,auditors}. These test whether a
reader is consistent, but cannot separate the two explanations, because a fingerprint that
survives the transformation predicts the same, correct answer.

The separating experiment needs the opposite edit: one that changes the function while
preserving specified within-layer summaries, the statistics a fingerprint can be built from. A reader relying on fingerprints should then answer as if
nothing had changed, while a reader that computes should follow the new function. We
construct such an edit, C2, and then ask when its outcome actually discriminates. The
second question turned out to be the more consequential one. Our contributions are:
\begin{enumerate}
\item \textbf{C2 for any linear readout.} Permuting the columns of the final linear layer
preserves every within-layer summary of every layer and can change the function in an exactly
recomputable way; every permutation we evaluate is selected and verified to change it (Section~\ref{sec:construction}).
\item \textbf{An eligibility protocol.} We state three conditions under which a C2 result
is informative about fingerprint use, plus a specificity check we learned we needed
(Section~\ref{sec:eligibility}).
\item \textbf{Negative eligibility results.} In four settings (tiny MLPs with a frontier
reader route, tiny MLPs with a self-hosted open-weight reader, and two public CNN zoos),
at least one condition fails in every case (Section~\ref{sec:applying}). In particular, the
frontier route's perfect C2 tracking is a feasibility result for the control, not a test of
fingerprint use.
\item \textbf{A positive control with an attested reader.} On a constructed population
that meets every condition, C2 separates a fingerprint reader from an executor, and a
self-hosted open-weight reader follows the rewired function rather than the available
shortcut. A registered replication on fresh networks confirms this, adds a second reader
family that also follows the rewiring, and confirms that without reasoning the reader does
not use the weights (Section~\ref{sec:planted}).
\item \textbf{An auditable artifact.} Exact transforms, a deterministic executor, frozen
prompt manifests, tool lockdown, free baselines, and script-generated audits.
\end{enumerate}

\section{Related work}
\paragraph{Language models as weight readers.} Weight-space programming showed frontier
models editing and describing small networks from weights~\citep{wsp}, and Weight
Oracles evaluates reading at larger scale~\citep{oracles}. Follow-up checks obfuscated
networks with hidden-unit shuffles and sign flips~\citep{huang}, permuted and sign-flipped
the hidden units of tabular clinical networks~\citep{auditors}, and audited neural-field
readers under neuron permutations and positive ReLU rescalings~\citep{orbit}. All of these
edits preserve the function. Nusrat and Nusrat plant known flaws in clinical decision
networks and compare weight reading with instructions that withhold the weights, a control
like our C4~\citep{nusrat}. Ahmed permutes output labels and scrambles values within
parameter arrays~\citep{memorization}, which respectively keep the learned weight
structure and destroy it. The wspbench repository scores forecasts of function-changing
edits in which a donor program with a different readout is recompiled~\citep{wspbench};
its donors are not matched to the original on layer statistics, so the edit is visible to
summaries alone. ObserverBench tests mechanistic estimates under intervention and
control~\citep{observer}. We did not find a column-only, spectrum-preserving readout edit
in these reports.

\paragraph{Weight-space learning.} A separate literature learns functions of weights
directly. Statistics of weights predict the accuracy of trained CNNs~\citep{unterthiner},
and model zoos provide populations for such studies~\citep{modelzoo}. Metanetworks and
neural functionals build the permutation symmetry of hidden units into the
architecture~\citep{dwsnet,nfn,gmn}; see \citet{wslsurvey} for a survey. Two points
connect to this work. First, these architectures are invariant to C1 by design but not to
C2, which changes cross-layer structure; C2 is therefore a natural stress test for them
as well as for language-model readers. Second, the fingerprint baselines in our
eligibility protocol are weight-space predictors of exactly this kind, and
\citet{unterthiner} show they can succeed. We evaluate them both with an entire
nonlinearity family held out and with their in-distribution splits
(Section~\ref{sec:applying}).

\paragraph{Faithfulness of reasoning.} Whether a reader's visible reasoning reflects its
computation is a question about chain-of-thought faithfulness~\citep{lanham}. Recent work
finds that reasoning is load-bearing mainly on harder tasks~\citep{loadbearing} and that
removing it returns models to baseline on serial tasks~\citep{ceiling}. Executing a
network's forward pass is such a serial task, which bears on our open-weight results.

\section{Construction}\label{sec:construction}
\begin{figure}[t]
\centering
\includegraphics[width=\linewidth]{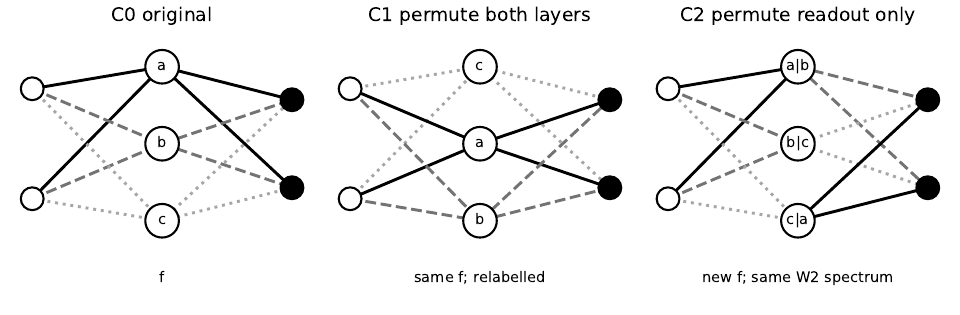}
\caption{The three weight arms. Edge style marks which original weight vector an edge
carries; node labels give the incoming row and outgoing column at each hidden slot. C1
moves both together and preserves the function. C2 permutes only the readout columns,
so each hidden unit now feeds the outputs another unit used to feed. A fourth arm, C4,
withholds all weight values.}
\label{fig:arms}
\end{figure}

Consider any network whose output is a linear readout of elementwise-activated features,
\[
f(x) = W\,\phi\big(g(x)\big) + b,\qquad W\in\mathbb{R}^{k\times m},
\]
where $g:\mathbb{R}^d\to\mathbb{R}^m$ is the rest of the network and $\phi$ acts
elementwise ($\phi$ may be the identity). Fully connected classifiers, and convolutional
classifiers whose last layer is linear, have this form.

\begin{proposition}\label{prop:c2}
Let $P$ be an $m\times m$ permutation matrix and $f'(x)=WP\,\phi(g(x))+b$. Then:
\begin{enumerate}
\item[(a)] Every parameter of $g$, and $b$, is unchanged. $WP$ has the same singular
values, Frobenius norm, entry multiset, and row norms as $W$, and its column norms are a
permutation of those of $W$. Hence any statistic $s$ of a single layer with
$s(WQ)=s(W)$ for all permutations $Q$ takes the same value on both networks.
\item[(b)] $f'(x)=W\phi\big(Pg(x)\big)+b$: C2 is the hidden-unit permutation of C1
applied to the incoming side only, without compensating the readout.
\item[(c)] $f'(x)=f(x)$ if and only if $W(P-I)\phi(g(x))=0$. On any finite input set,
the inputs whose output changes are therefore found exactly by one forward pass of each
network.
\end{enumerate}
\end{proposition}

\begin{proof}
(a) $P$ is orthogonal, so $(WP)(WP)^\top=WW^\top$, which fixes the singular values and
the Frobenius norm. Right multiplication by $P$ reorders the columns of $W$, which fixes
the entry multiset and each row's entries and permutes the columns. The invariance of $s$
is the definition with $Q=P$. (b) An elementwise $\phi$ commutes with coordinate
permutations, so $P\phi(z)=\phi(Pz)$. (c) Subtract the two expressions.
\end{proof}

Proposition~\ref{prop:c2}(a) covers within-layer summaries only. Cross-layer statistics
can change: C2 alters which incoming weight row is paired with which readout column, and
we report such a pairing statistic below. C2 is a transparent counterfactual, not an
information-theoretic obfuscation. Part (c) also shows why $P$ must be selected: a
permutation can leave $f$ unchanged, for example when it swaps dead or duplicate units.

\paragraph{Tiny MLPs.} Our main reader study uses $f(x)=W_2\operatorname{ReLU}(W_1x+b_1)+b_2$
with four inputs, eight hidden units, and three classes, so $g(x)=W_1x+b_1$ and $W=W_2$.
Readers see rounded values, and the executor scores against the same rounded
serialization. C1 permutes hidden units consistently in both layers. For C2, derangements
are selected on a disjoint selection set and scored on a held-out scoring set; every
included network changed its prediction on 10 to 16 of the 16 scoring inputs.
Inert permutations (those that change no prediction) never enter a reported score, since
readers are scored only on inputs where C0 and the selected C2 disagree. They are also
rare here: none of 64 random derangements per network was inert on the scoring inputs, and
the selected C2 changed the prediction on a median 86.9\% of 10,000 random inputs, moving the
logits by a median 1.65 times the mean C0 logit norm. In the planted-pairing control of
Section~\ref{sec:planted}, a swap of two units that vote for the same class is often inert
on the scoring inputs (median 23\% of all 28 single swaps in the registered population), which is why
C2 is restricted to units with different preferred classes; among those, the median inert
share is 0.

\paragraph{CNN zoos.} In the model-zoo preflights, C2 permutes the columns of each
network's final linear layer, and C1 permutes the same units consistently in both
adjoining layers (Section~\ref{sec:applying}).

\section{When does C2 discriminate? An eligibility protocol}\label{sec:eligibility}
A C2 experiment compares two hypotheses. A fingerprint reader maps C2-invariant summaries
to an answer, so by Proposition~\ref{prop:c2}(a) it must answer C2 as it answered C0. A
computing reader follows $f'$. The experiment discriminates only if both hypotheses are
live and make different, checkable predictions. This gives three conditions.

\begin{description}
\item[E1. A fingerprint strategy works on C0.] A predictor restricted to C2-invariant
summaries, evaluated on held-out networks, predicts the target on C0 better than a
constant by a preset margin. Otherwise the evaluated invariant-summary baselines do not explain C0
success, and C2 cannot separate the hypotheses on this target. E1 is an operational
criterion: it does not rule out stronger invariant strategies. We required at least
a 10\% lower mean absolute error (MAE) than the training mean, with whole families held
out.
\item[E2. C2 is mechanically sound.] C1 reproduces C0 outputs and the C2 invariants hold
to numerical precision, and C2 moves the target in most networks. Our preset thresholds
were: maximum C1 logit error and maximum spectrum error at most $10^{-5}$; at least 80\%
of networks moving accuracy by at least 0.5 points; at least 48 networks in at least four
families.
\item[E3. The reader is attested.] The model that produced each answer is identified,
for example open weights at a pinned revision, or an interface that reports the resolved
backend.
\end{description}

\paragraph{Specificity (added post hoc).} The CIFAR-10 preflight showed that E2 is not
enough. If a generic random readout swap predicts the C2 effect on the target, a reader
that matches C2 accuracy need not have followed the specific rewiring. We therefore
recommend a fourth check, E2s: the target should be specific to the chosen permutation,
for example per-input outputs rather than aggregate accuracy, or should be shown not to be
predictable from a random swap. We defined this check after seeing the CIFAR-10
accuracy result; its per-image version was then run under a protocol committed before
the job (Section~\ref{sec:applying}).

E1 and E2s must hold for the \emph{same} target. A target on which a fingerprint strategy
works but C2's effect is generic, or one on which C2's effect is specific but no
fingerprint strategy works, cannot discriminate.

\section{Applying the protocol}\label{sec:applying}
Table~\ref{tab:eligibility} summarises the four settings. The rest of this section gives
the evidence for each entry.

\begin{table}[t]
\centering
\small
\caption{Eligibility of each setting. Pass/fail refer to the conditions of
Section~\ref{sec:eligibility}; n.a.\ means the condition does not apply because no
reader was run. $^\ast$In-distribution only; E1 fails when a whole nonlinearity family is
held out. $^\dagger$A reader run was prepared and gated on E1, which failed. T$-$S: tracking minus stuck. No natural
setting meets all conditions. The last row is a constructed positive control
(Section~\ref{sec:planted}).}
\label{tab:eligibility}
\begin{tabular}{@{}lccccl@{}}
\toprule
Setting (networks) & E1 & E2 & E2s & E3 & Reader outcome \\
\midrule
Tiny MLPs, frontier route (24) & fail & pass & pass & fail & C0 = C1 = C2 = 1.000 \\
Tiny MLPs, Qwen3.8-27B (24) & fail & pass & pass & pass & capability gate failed \\
MNIST CNN zoo (60) & fail & pass & not run & n.a. & no reader \\
CIFAR-10 zoo, accuracy (60) & pass$^\ast$ & pass & fail & n.a. & no reader \\
CIFAR-10 zoo, per image (60) & fail & pass & pass & n.a. & reader not run$^\dagger$ \\
\midrule
Planted pairing, synthetic (24) & pass & pass & pass & pass & Qwen: C2 track 0.906, stuck 0.010 \\
Planted pairing, registered (24) & pass & pass & pass & pass & T$-$S: Qwen 0.875, DeepSeek 0.729 \\
\bottomrule
\end{tabular}
\end{table}

\subsection{Tiny MLPs with a frontier reader route}\label{sec:mlp}
On 24 held-out tiny MLPs, a frontier model reached through a command-line subscription
interface (requested label GPT-6.1 Sol, tools disabled) scored 1.000 on C0, C1 and C2 and
0.292 with weights withheld, and followed the rewired function on every input where C0
and C2 disagree (24/24 networks; exact 95\% interval for the share $[0.858, 1]$). Across a
size ladder from 67 to 4,099 parameters it stayed at ceiling while its output tokens grew
with network size (slope \ladderslope{} on log scales). Two conditions fail, so this is a
feasibility result for the control, not a test of fingerprint use: invariant-summary
baselines are at chance on C0 (E1), and the backend cannot be attested (E3).
Appendix~\ref{app:frontier} gives the protocol, statistics, baselines and ladder.

\subsection{Tiny MLPs with an open-weight reader}
To satisfy E3 we self-hosted Qwen3.8-27B at a pinned revision on the same 24 networks
and 72 prompts (C0, C2, C4), under a gate protocol fixed before any output existed: mean
C0 accuracy at least 0.60, the lower bound of the 95\% bootstrap interval for C0 minus C4
above 0.20, and at most 10\% parse failures. With an empty reasoning block, the model
scored 0.297 on C0 and 0.292 on C4, a difference of $+0.005$ (95\% CI $[-0.107, 0.117]$,
24 networks), consistent with reports that removing reasoning returns models to baseline
on serial tasks~\citep{ceiling}. With reasoning enabled, 71 of 72 trials were still
computing when they reached the preset 8,192-token cap, so the run failed its gate. The
cap was set too low; the failure shows that the model did not finish within the budget,
not that it cannot.

\subsection{Model-zoo preflights}
C2 needs a setting where E1 holds, so we looked for one offline, with no reader, in two
public populations of trained CNNs~\citep{modelzoo}: 60 MNIST and 60 CIFAR-10
networks, each spanning four nonlinearity families (GELU, ReLU, sigmoid, tanh). The
fingerprint predictor was a ridge regressor on per-layer norms, quantiles, and singular
values, with C0 test accuracy as the target. The gate requires a mean absolute error
(MAE) at least 10\% below that of the training-set mean.

\paragraph{Held-out families: E1 fails in both.} Evaluated leave-one-family-out, the
regressor's MAE was 0.0215 against 0.0157 for the training mean on MNIST, and 0.194
against 0.153 on CIFAR-10. An earlier frozen MNIST analysis had used the C2 effect rather
than C0 accuracy as the target; correcting the target did not change the outcome (MAE
0.0254 against 0.0157 with the original features).

\paragraph{In-distribution: E1 holds on CIFAR-10.} \citet{unterthiner} predicted
accuracy well from similar statistics with in-distribution splits, so we ran that test
too, under a protocol committed before the job ran: 20 repeats of random 5-fold
cross-validation over the same 60 networks, with the same features, penalty, and 10\%
rule. To separate fingerprints from mere recognition of the nonlinearity family, which
is easy to read off weights and shifts mean accuracy, the protocol also fixed a
family-mean baseline that predicts each held-out network by its family's training-fold
mean. On CIFAR-10 the regressor's MAE was 0.0748 (range over repeats 0.0659--0.0861),
against 0.1218 for the training mean and 0.0861 for the family mean: the gate passes,
and the regressor also beats the family-mean baseline by more than 10\%. On MNIST the
regressor did worse than a constant (0.0159 against 0.0132; family mean 0.0116), so
family membership is the only usable signal there. CIFAR-10 is therefore the first
setting we found in which a fingerprint strategy works on C0, though only for networks
from families seen in training.

\paragraph{E2 passes, E2s fails on CIFAR-10.} On CIFAR-10, C1 reproduced C0 logits to
within $4\times10^{-15}$ over the full test set in float64, the readout spectrum was
preserved to the same precision, and C2 moved accuracy by at least half a point in 49
of 60 networks. The MNIST population also passed the mechanics checks. However, on
CIFAR-10 a random swap of two readout columns predicted each network's C2 accuracy with
correlation 0.975. With accuracy as the target, a reader matching C2 accuracy on these
networks need not have followed the specific rewiring. The CIFAR-10 zoo thus satisfies
E1 (in-distribution) and E2 but not E2s.

\paragraph{Per-image targets: E2s passes, E1 fails.} We then changed the target to each
network's predicted class on individual test images, under a protocol committed before
either job. For specificity, we compared C2 with eight random two-column readout swaps
per network on the 10,000 test images. In the typical network a random swap reproduced
13.8\% of C2's changed predictions (median over networks; quartiles 10.6--21.2\%,
maximum 28.7\%), and 52 of 60 networks had at least 50 changed images (median 1,478).
C2 changed no prediction in 8 networks, 7 of them sigmoid. Per image, C2's effect is
therefore specific to its column pair. For E1 we gave two predictors only what C2 leaves
unchanged in the final layer: the penultimate activations $h$, the bias, and the
readout up to a column permutation. One scales the readout's row means by $\sum_j h_j$;
the other pairs the largest activations with the largest-norm columns. On rounded
values over all 60 networks they agreed with the C0 class on 22.0\% and 21.5\% of
images, below the 22.9\% obtained by always predicting a network's most frequent class
(the gate required 32.9\%). We had prepared an attested-reader run on this target, with
Qwen3.8-27B at the pinned revision, a 32,768-token cap, and 24 networks with one changed
image each. The protocol made it conditional on E1, so it was not run. On this zoo, the
two conditions trade off: accuracy admits a fingerprint strategy but not a specific
effect, and per-image classes admit a specific effect but no fingerprint strategy we
tested.

\subsection{A positive control: planted pairing}\label{sec:planted}
No natural setting met E1 and E2s on the same target, so we built one, to check that C2
does discriminate when the conditions hold. The networks have the main study's shape
(4 inputs, 8 hidden units, 3 classes) with a planted rule that can be read off a single
network: hidden unit $j$ votes for the class $k_j=\arg\max_{c<3}W_1[j,c]$, so
$W_2[c,j]=a_j[c=k_j]+0.1\,\epsilon$ with $a_j\sim U(1,2)$ and $\epsilon\sim N(0,1)$. C2
swaps the readout columns of two units with different preferred classes, chosen on a
selection set, and a network is kept if C2 changes at least 4 of its 16 scoring
predictions. Two readers with known strategies serve as references. The executor X
runs the forward pass. The fingerprint reader F applies the rule using only what C2
leaves unchanged:
$\mathrm{logits}_c=b_{2,c}+\bar a\sum_j \operatorname{ReLU}(W_1x+b_1)_j\,[c=k_j]$, where
$\bar a$ is the mean of the column maxima of $W_2$, so by
Proposition~\ref{prop:c2}(a) F repeats its C0 answers on C2.

\paragraph{A failed first criterion, and a fresh population.} The first protocol
required F's tracking-minus-stuck on C2-changed inputs to be at most $-0.90$. On its 24
networks E1 passed (F accuracy 0.938 against 0.508 for the majority class) and E2s
passed (a random swap reproduced a median 19.8\% of C2's changes), but F scored $-0.872$,
so that stage failed and no reader was run. The criterion was miscalibrated: because F
repeats its C0 answers, its stuck rate on changed inputs equals its C0 accuracy there,
so the bar demanded about 95\% accuracy rather than separation. We did not revise the
threshold on those data. A second protocol, committed before any new network existed,
replaced the criterion with two conditions (F predicts the C2 class on at most 10\% of
changed inputs, and X's tracking-minus-stuck exceeds F's by at least 1.5) and drew 24 fresh
networks with separate seeds.

\paragraph{C2 separates the two readers.} On the fresh population E1 passed (F accuracy
0.948 against 0.602), E2s passed (median 19.0\%), and F predicted the C2 class on 5.7\%
of changed inputs while X tracked all of them: a separation of 1.856 on the scale from
$-2$ to $+2$. Here C0 accuracy alone cannot tell the two readers apart (0.948 against
1.000), but C2 can. Each network had 4 to 11 changed scoring inputs (median 6).

\paragraph{An attested reader.} \readerresult

\paragraph{The same reader without room to compute.} To see whether C2 would catch this
reader using a shortcut, a further protocol, committed before its run, repeated the 72
prompts with thinking disabled and a 64-token cap. We found no evidence that the reader
used the weights: C0 accuracy 0.323 against 0.385 with weights withheld (difference $-0.063$,
[$-0.219$, 0.094]), so the gate (lower bound above 0.10) failed and C2 had nothing to
discriminate. Under this configuration we found no evidence of weight use; with reasoning, its outputs
follow the rewired function. No C2-invariant strategy appeared in it to be caught, so the
scripted reader F remains our only demonstration of C2 detecting fingerprint use.

\paragraph{A registered replication with a second reader family.} Amendments 06 and 07
were exploratory. We therefore registered a replication (\ifanon
\else AsPredicted \#323988, dated 2026-11-11 and made public with this preprint; the
registration PDF is available from the authors\fi) before generating its networks:
the same design on 24 fresh networks from new seeds, with two confirmatory hypotheses and
two secondary readers. All runs used one source commit and the pinned revisions in
Appendix~\ref{app:repro}. Stage 1 passed again (F accuracy 0.935 against 0.570 for the
majority class; median reproduced share 0.177; F predicted the C2 class on 7.4\% of changed
inputs; separation 1.848). \emph{H1}, that Qwen3.8-27B in think mode passes the gate and
tracks C2, was supported: C0 accuracy 1.000 against 0.406 with weights withheld (difference
0.594 [0.510, 0.677]), 3 of 72 answers unparseable, and on C2 tracking 0.875 with stuck
0.000, tracking minus stuck 0.875 [0.750, 1.000], so even the lower bound exceeds 0.5.
\emph{H2}, that the same reader with thinking off fails the gate, was supported: C0 and C4
accuracy were both 0.417 (difference 0.000 [$-0.156$, 0.156]). Among the secondary readers,
DeepSeek-R1-Distill-Qwen-14B passed the gate (0.427 [0.271, 0.573]; 6 of 72 unparseable) and
tracked C2: tracking 0.771, stuck 0.042, tracking minus stuck 0.729 [0.552, 0.885].
gpt-oss-20b at high reasoning effort failed the gate: 51 of its 72 prompts reached the
32,768-token cap before any final answer, so the cap, not the control, limited it. The outputs of two
reasoning readers from different families thus track the rewired function on networks
where a C2-invariant shortcut scores 0.935. For each reader we used the first of two
copies to finish; the second copies, on other hardware, gave the same readings, but the
two DeepSeek copies agreed on only 38 of 72 answers: greedy decoding over roughly 11,000
reasoning tokens in bfloat16 is not reproducible across GPU types, though the network-level
conclusion was.

\section{Limitations}
Tool lockdown cannot rule out mental execution, so on tiny networks a computing reader
and a reader that simulates code are the same thing; this is intended, but it means the
study says nothing about shortcuts that are faster than execution. The protocol's
conditions are necessary for a discriminating C2 test, not sufficient: Proposition
\ref{prop:c2} covers within-layer summaries, and a reader could exploit cross-layer
statistics that C2 changes. The frontier route's results rest on one requested label
with an unattested backend. Every result except the registered replication of
Section~\ref{sec:planted} is exploratory: those protocols were committed to the repository
before their results but not registered externally, and the specificity check was added
after seeing the CIFAR-10 data. The per-image E1 test used two simple invariant predictors; a stronger invariant
strategy might succeed where they failed. The positive control uses two reasoning readers
(and one that hit its token cap) on constructed networks with a single planted rule; a
reader could compute here and still use fingerprints on trained networks. Long greedy
reasoning runs did not reproduce answer by answer across GPU types, so per-prompt results
should be read through the network-level intervals. We have not observed an LLM using a C2-invariant shortcut:
without reasoning, we found no evidence that the attested reader used the weights, so only the scripted
reader F shows C2 detecting fingerprint use. Sample sizes are small (24 reader networks,
16 ladder networks, 60 networks per zoo).

\section{Conclusion}
C2 is a simple, general, exactly checkable control: permute a readout's columns, and every
within-layer summary stays fixed while the function can change (we verify that it does). Using it to test whether
language models read weights through fingerprints requires a setting in which a
fingerprint strategy works, the edit changes a specific recomputed property, and the
reader is attested. We found no setting that meets all three. The closest is the
CIFAR-10 zoo, where the conditions traded off: weight statistics predict accuracy
in-distribution, but C2's effect on accuracy is generic; C2's effect on per-image
predictions is specific, but no invariant strategy we tested predicts those
predictions. When the conditions are built in, C2 works: on a planted-pairing population
it separates a rule-following reader from an executor, and in a registered replication the outputs of two
attested open-weight readers from different families tracked the rewired function even
though the shortcut was available. Output tracking does not identify the readers' internal
strategy. The open problem is a
natural target that is both fingerprint-predictable and
specific to the rewiring, for example a per-image target on a population where trained
readouts share structure, or a stronger invariant predictor. Only then is an attested
reader run worth its cost.

\ifanon\else
\section*{Acknowledgements}
Computing resources were provided by the Center for Computational Research at the
University at Buffalo.
\fi

\section*{Reproducibility statement}
\ifanon
\else
On request, the authors provide the code (transforms, audits, reader harness and job
scripts), prompt manifests with SHA-256 hashes, raw reader records and the registration PDF.
Nothing is embargoed beyond this version; a public repository with an archived DOI will
accompany the next version of this preprint.
\fi
Appendix~\ref{app:repro} lists the models, revisions, dates, and checks.

\bibliography{refs}
\bibliographystyle{tmlr}

\clearpage
\appendix
\section{Exact prompt}\label{app:prompt}
Every tiny-MLP prompt follows this template, produced by \texttt{bwf/prompts.py}; the
values in angle brackets are filled from the manifest. C4 prompts add the line marked
\texttt{[C4]} and carry weight shapes without values.
\begin{small}
\begin{verbatim}
The network is a one-hidden-layer ReLU classifier: logits = W2 @ relu(W1 @ x + b1) + b2.
For every input, predict its class. Then, for each requested hidden unit,
predict every class after setting that unit's W2 column to zero.
Reply with JSON only, exactly this schema:
{"predictions":[0,2,1],"ablation":{"<u>":[0,2,1]}}.
[C4] Weights are withheld. You must still provide your best-guess class for every
input; do not refuse or leave any prediction blank.
Do not add markdown or explanation.
DATA={"network":<serialized weights>,"inputs":<16 inputs>,"ablation_units":[<u>]}
\end{verbatim}
\end{small}
Line breaks inside the first three sentences are added here for width. Replies are
parsed by a strict parser that accepts only this schema.

\section{Models, dates, and audits}\label{app:repro}
\begin{itemize}
\item \textbf{Frontier route}: requested labels GPT-6.1 Sol (analysed) and GPT-6 Astra,
through Codex CLI 0.167.0-alpha.3.2 in read-only ephemeral mode with tools disabled;
96 trials each, run 2026-11-07. Astra's run was paused after 12 trials and resumed
with a retry-and-timeout-only change (commits \texttt{700e0b0} to \texttt{f784186}); the
resume reused a trial only when its prompt hash matched.
\item \textbf{Size ladder}: same labels through Codex CLI 0.166.0-alpha.6, 64 trials per
label, recorded 2026-11-05.
\item \textbf{Open-weight reader}: \texttt{Qwen/Qwen3.8-27B} at revision
\texttt{37561816e6}, greedy decoding, recorded 2026-11-09; a secondary
\texttt{DeepSeek-R1-Distill-Qwen-14B} at revision \texttt{1df8507178} is descriptive only.
Prompt-manifest SHA-256 \texttt{0266214014fae0cd\ldots ab21}.
\item \textbf{Registered replication (Amendment 08)}: \texttt{Qwen/Qwen3.8-27B} at
\texttt{37561816e6} (think, 32,768-token cap; and thinking off, 64-token cap),
\texttt{openai/gpt-oss-20b} at \texttt{6cee5e81ee} (reasoning effort high; answer taken
from the final channel), and \texttt{deepseek-ai/DeepSeek-R1-Distill-Qwen-14B} at
\texttt{1df8507178}; bfloat16, greedy, batch 4, one H100 PCIe per job, source commit
\texttt{8fccf5f}; 72 prompts per reader, recorded 2026-11-12. Prompt-manifest SHA-256
\texttt{fa08a13a9b6b0682\ldots d6d3cd}.
\item \textbf{Model zoos}: 60 MNIST and 60 CIFAR-10 CNN checkpoints from
\citet{modelzoo} (MNIST: 15 per nonlinearity family); offline analyses recorded 2026-11-08
and 2026-11-09.
\end{itemize}
Four script-generated audits accompany the reader results. The replication audit checks
the 96 analysed prompt hashes against the manifest and recomputes the bootstrap
intervals. The baseline audit rebuilds the rounded C0, C1, and C2 networks from the
stored prompts and evaluates the free baselines. The validity audit checks completeness,
finite values, C0/C2 disagreement, and that prompt data contain no labels or answers. The
forensic audit compares the two requested labels' records, documenting identical final
answers and distinct invocation provenance. Raw records keep responses, event streams,
tool-event counts, executable versions, prompt hashes, and token accounting.

\section{Frontier reader route: full results}\label{app:frontier}
\paragraph{Protocol.} We evaluated 24 held-out networks (seeds 10000--10023) in four
arms (C0, C1, C2, C4) with one requested model label, GPT-6.1 Sol, accessed through a
command-line subscription interface: 96 trials. C4 replaces weight values with shape
metadata and requires a best guess. Each prompt asks for the class of 16 inputs and for
the classes after zeroing the readout column of one seeded active hidden unit
(Appendix~\ref{app:prompt}). Prompts were fixed in an offline manifest before any reader
call. The run is exploratory. Calls used fresh, ephemeral, read-only sessions with shell, web,
browser, MCP, and plugin tools disabled. A command-execution canary ran before each
batch and every ten trials, and any non-message tool event invalidated its trial.

\paragraph{Statistics.} The base network is the unit of analysis. Intervals are 95\%
percentile intervals from 10,000 paired bootstrap resamples of the 24 networks. The
primary contrasts are C0 minus C4 accuracy, and, on the inputs where C0 and C2
disagree, the rate of tracking $f'$ minus the rate of sticking to $f$.

\begin{table}[t]
\centering
\small
\caption{Frontier reader route on 24 tiny MLPs (16 inputs each, 96 trials). Accuracy is
the mean over networks. The bootstrap interval for tracking minus stuck collapses to
$[1,1]$ because every network scored $+1$; the exact interval is in the text.}
\label{tab:results}
\begin{tabular}{@{}lcccc@{}}
\toprule
& C0 & C1 & C2 & C4 \\
\midrule
Accuracy & 1.000 & 1.000 & 1.000 & 0.292 \\
\midrule
C0 $-$ C4 accuracy & \multicolumn{4}{l}{$+0.708$, 95\% CI $[+0.638, +0.779]$} \\
C2 tracking $-$ stuck & \multicolumn{4}{l}{$+1.000$; 24/24 networks fully tracked,
exact 95\% CI for the share $[0.858, 1]$} \\
\bottomrule
\end{tabular}
\end{table}

\begin{figure}[t]
\centering
\includegraphics[width=.8\linewidth]{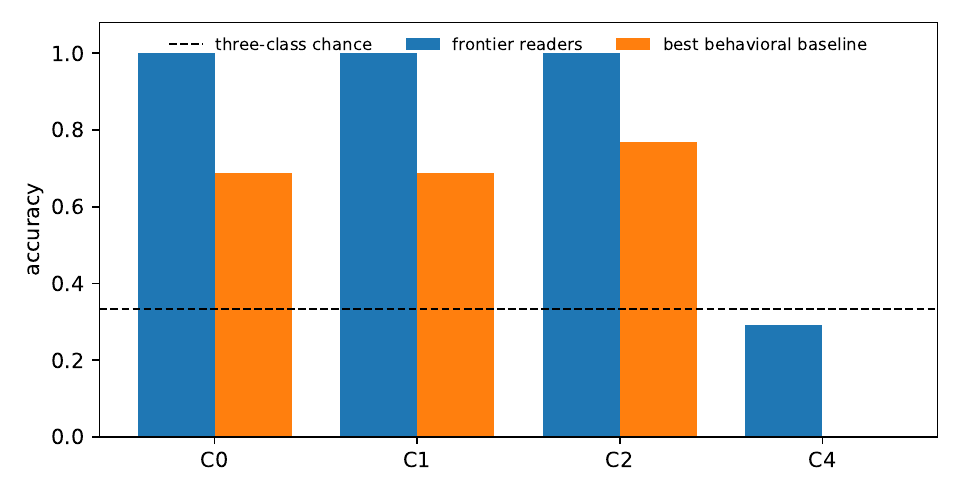}
\caption{Reader accuracy and the strongest behavioural baseline (template matcher) on
the 24 tiny MLPs. The C4 baseline bar is omitted because no weights are available to it.}
\label{fig:results}
\end{figure}

\paragraph{Results.} Table~\ref{tab:results} and Figure~\ref{fig:results} give the
results. The route followed the rewired function on every C0/C2 disagreement input of
every network. Because each network's tracking-minus-stuck value is exactly $+1$, every
bootstrap resample has mean $+1$ and the percentile interval is degenerate. A more
informative statement treats each network as a success or failure: 24 of 24 networks
were tracked on all disagreement inputs, and the exact (Clopper--Pearson) 95\% interval
for the population share is $[0.858, 1]$. C1 success is a pipeline check, and C4 gives
the floor without weights. There were no parse failures and no tool-invalidated trials
(0/96 each).

\paragraph{Free baselines.} Table~\ref{tab:baselines} lists baselines that need no reader.
The two that use only summaries a fingerprint reader could exploit, a layer-statistics
regressor and a spectral predictor, both trained leave-one-orbit-out, score 0.294 and
0.258 on C0, at or below the three-class chance level of 1/3. This is why E1 fails. The
strongest baseline, a linear template matcher that does use the weight values, scores
0.688 on C0 and 0.768 on C2. A cross-layer pairing statistic, the agreement between
$W_1$ row norms and $W_2$ column norms, is higher for C0 than for C2 in 19 of 24 networks
(mean difference $+0.524$), so C2 is detectable through cross-layer structure, as
Proposition~\ref{prop:c2} allows, although that statistic does not predict outputs.

\begin{table}[t]
\centering
\small
\caption{Free baselines on the 24 tiny MLPs (accuracy over 16 inputs per network).}
\label{tab:baselines}
\begin{tabular}{@{}lccc@{}}
\toprule
Baseline & C0 & C1 & C2 \\
\midrule
Linear template matcher & 0.688 & 0.688 & 0.768 \\
Output-row norm balance & 0.297 & 0.299 & 0.424 \\
Layer-statistics regressor, leave-one-orbit-out & 0.294 & 0.344 & 0.372 \\
Spectral predictor, leave-one-orbit-out & 0.258 & 0.263 & 0.349 \\
\bottomrule
\end{tabular}
\end{table}

\paragraph{Why this is not a test of fingerprint use.} Since no invariant-summary
baseline approaches the reader on C0, the evaluated invariant-summary baselines do not account for the
reader's C0 accuracy. The route's outputs tracked the rewired function on the evaluated
inputs, which establishes that the control works as
intended with a capable reader; it does not discriminate between hypotheses. Two further
limits apply. Each C2 prompt is fresh and contains no C0 weights for a reader to remain
attached to. And E3 fails: a second requested label (GPT-6 Astra), run on the same
prompts, produced final JSON identical to the analysed route on every prompt, with
different event streams, usage records, and thread identifiers. The interface does not
report the resolved backend, so we treat it as provenance of separate invocations, not as
an independent replication.

\begin{figure}[t]
\centering
\includegraphics[width=.6\linewidth]{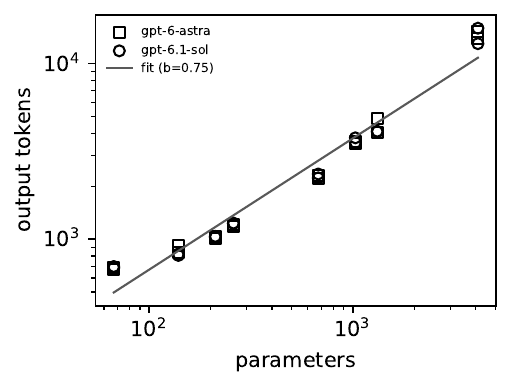}
\caption{Output tokens per trial (interface usage record, mean over the four arms)
against parameter count for the 16 ladder networks, one marker per network and route,
with the least-squares fit on log scales.}
\label{fig:ladder}
\end{figure}

\paragraph{Size ladder (exploratory).} Eight rungs spanned depths 1--3 and widths 8--512
(67 to 4,099 parameters), with two networks per rung (16 networks) and the same four
arms. Both requested labels scored 1.000 on C0, C1, and C2 at every rung, with C2
tracking of 1.000; mean C4 accuracy was 0.367. Output tokens in the interface's usage
record, almost all of which are not part of the final JSON answer, rose for the analysed
route from about 690 per trial at 67 parameters to about 14,400 at 4,099
(Figure~\ref{fig:ladder}). A least-squares fit of log tokens on log parameters, over the
32 network-and-label observations, gives slope \ladderslope. The final JSON answers
themselves stayed at about 25 tokens at every size. Ceiling accuracy together with generated work that grows with network size
fits serial execution of the forward pass better than fingerprint recognition. With two
networks per rung, per-rung accuracy has no useful interval.

\section{Protocol history}\label{app:history}
Every analysis after the main reader run followed a written protocol committed to the
repository before its job ran. Table~\ref{tab:history} lists them in order with their
outcomes, including the failed gates and the one recorded job failure.

\begin{table}[h]
\centering
\small
\caption{Protocols and outcomes. Each protocol was committed to the repository before its
results existed; all are exploratory except Amendment 08, which was registered externally
before its networks were generated.}
\label{tab:history}
\begin{tabular}{@{}p{0.27\linewidth}p{0.33\linewidth}p{0.32\linewidth}@{}}
\toprule
Protocol & Question & Outcome \\
\midrule
Amendment 01 & Zoo mechanics and held-out-family fingerprint gate & Mechanics passed (CIFAR-10); fingerprint gate failed on MNIST and CIFAR-10 \\
Open-reader gate & Can Qwen3.8-27B do the tiny-MLP task at 8,192 tokens? & Failed; 71/72 trials at the token cap \\
Amendment 02 & In-distribution fingerprint gate & Passed on CIFAR-10, failed on MNIST \\
Amendment 03 & Per-image specificity on CIFAR-10 & Passed \\
Amendment 04 & Per-image fingerprint gate, then a reader & Gate failed; reader not run. One crash before any output, fixed and recorded \\
Amendment 05 & Planted-pairing positive control & E1, E2s passed; discrimination criterion failed (miscalibrated) \\
Amendment 06 & Corrected criterion, fresh population, then Qwen at 32,768 tokens & Stage 1 passed; Qwen gate passed, C2 tracking minus stuck 0.896 \\
Amendment 07 & Same Qwen prompts, thinking off, 64-token cap & Gate failed; no evidence of weight use under this configuration \\
Amendment 08 (registered) & Replication on fresh networks; Qwen think and no-think; gpt-oss-20b and DeepSeek-R1-14B & Stage 1 passed; H1 and H2 supported; DeepSeek tracks (0.729); gpt-oss gate failed at the token cap \\
\bottomrule
\end{tabular}
\end{table}

\end{document}